\documentclass[letterpaper, 10 pt, conference]{ieeeconf}  

\IEEEoverridecommandlockouts                              

\usepackage{graphicx}
\usepackage{subfig}
\usepackage{cite}
\usepackage{multicol}

\usepackage{algorithm}
\usepackage{algorithmic}
\usepackage{color}
\usepackage{eqparbox,array}
\usepackage{wrapfig}

\usepackage{float}
\newfloat{algorithm}{t}{lop}

\usepackage{amsmath,amssymb}
\DeclareMathOperator*{\argmin}{arg\,min}
 
\let\vec\boldvec

\newlength{\defaulttextfloatsep}
\newlength{\defaultintextsep}
\usepackage{hyperref}

\title{\LARGE \bf
Uncertainty Aware Learning from Demonstrations in Multiple Contexts using Bayesian Neural Networks}

\author{Sanjay Thakur$^{1}$, Herke van Hoof$^{2}$, Juan Camilo Gamboa Higuera$^{1}$, Doina Precup$^{1}$, and David Meger$^{1}$
\thanks{$^{1}$School of Computer Science, McGill University}%
\thanks{$^{2}$Informatics Institute, University of Amsterdam}
}

\begin{document}

\maketitle
\thispagestyle{empty}
\pagestyle{empty}

\begin{abstract}
Diversity of environments is a key challenge that causes learned robotic controllers to fail due to the discrepancies between the training and evaluation conditions. Training from demonstrations in various conditions can mitigate---but not completely prevent---such failures. Learned controllers such as neural networks typically do not have a notion of \emph{uncertainty} that allows to diagnose an offset between training and testing conditions, and potentially intervene. In this work, we propose to use Bayesian Neural Networks, which have such a notion of uncertainty. We show that uncertainty can be leveraged to consistently detect situations in high-dimensional simulated and real robotic domains in which the performance of the learned controller would be sub-par. Also, we show that such an uncertainty based solution allows making an informed decision about when to invoke a fallback strategy. One fallback strategy is to request more data. We empirically show that providing data only when requested results in increased data-efficiency.
\end{abstract}

\section{INTRODUCTION}
This paper addresses the problem of learning from demonstration (LfD) when the environment changes frequently and outside the control of the robot. This situation is common in real-world deployments, such as self driving cars that must operate at all times of the day, in many traffic conditions and in all seasons. The performance of an LfD agent on an environment is dependent on the coverage and diversity of previous demonstrations. Training over multiple conditions has been shown to result in controllers that are more robust to such situations~\cite{rajeswaran2016epopt}. Furthermore, by making the policy dependent on the history, policies can rapidly adapt to different conditions~\cite{wang2016learning, RL2}. These robustness methods work up to a point, but cannot be expected to accommodate drastic and unexpected changes in the environment. As such, we think the agent should be able to signal conditions where it is not confident, allowing it to switch to a fall-back strategy.

We propose using Bayesian Neural Networks~(BNNs) \cite{MCDropout,BayesByBackprop} to learn robotic control policies from demonstration to model complex control functions while simultaneously assessing the uncertainty in their own predictions. We consider the case where task variables are not directly observed. For example, we may need to take several strides in the morning before determining that a joint is stiff today and will require additional force for optimal gait. Therefore, we train our networks on temporal windows of experience and observe that this allows effective assessment of uncertainty across different variations of a task which we will call \emph{contexts}. In this setup, high predictive uncertainty indicates that the robot is currently encountering a context on which the controller has not been sufficiently trained. A low predictive uncertainty, on the other hand, reflects high confidence that the learned policy will be successful in the current context. Our contributions in this work are 
\begin{itemize}
    \item to design a framework to generate predictive uncertainties over multiple task contexts as a confidence measure in an LfD setting scalably, and
    \item to propose an approach for actively requesting training data where the controller is not confident.
\end{itemize}

We support our claims through our experiments on a real and simulated robots. We first show that BNNs scale up easily to higher-dimensional tasks. We then show that there exists an inverse relation between the predictive uncertainty given by the BNN and the performance of the controller, so, the uncertainty can be used to predict its success. Finally, as fallback strategy we propose requesting more demonstrations when the predictive uncertainty is too high. We show that seeking more context specific demonstrations only when the context is highly uncertain leads to data-efficient learning.

In the next section, we define the objectives of our approach. We then describe related prior work such as active LfD and learning across multiple task variants. Next, we formalize our problem set-up and discuss relevant background. Subsequently, we outline how we use BNNs to represent uncertainty and how we leverage the uncertainty in the context of active LfD on multiple task contexts. We demonstrate the viability and benefits of our approach in section~\ref{section:Results} on a physical pendulum swing-up task, as well as several simulated robotics tasks.

\section{Objectives and Notation}\label{section:objectives_and_notation}
We denote by $D$ the set of task contexts that a robot might encounter one after another sequentially with each context $d$ defined through a reward function $R_d({\bf s}_t,{\bf a}_t)$ and state transition dynamics $T_d({\bf s}_t,{\bf a}_t)$ which take as arguments an action ${\bf a}_t$ and a state ${\bf s}_t$. Each context $d$ has a demonstrator $\pi^*_d$ which provides demonstrations on $d$ upon request, which are merged with the complete dataset $\mathcal{D}$. This demonstrator is (approximately) optimal with respect to the cumulative reward obtained on context $d$ over the horizon $H$ number of steps, $\sum_{t=1}^H R_d({\bf s}_t, {\bf a}_t)$. 

As an example of a fallback strategy, we maintain a set-up where for each context, the controller which is being trained can halt execution at any point and ask for more demonstrations. The objective of the controller is to maximize the overall cumulative reward obtained on all the contexts in $D$ while limiting the number of requests for demonstrations. In order to do so, the controller should have the ability to predict whether it will be able to succeed at the current context without additional demonstrations.

\section{RELATED WORK}\label{section:Related_Work}
Our work aims to develop a method that allows a controller trained on demonstrations in several contexts to predict in which new contexts it might perform poorly. If this were the case, we propose that the learner should actively ask for more demonstrations under the current context.
In this section, we will thus cover some related work on the topics of active learning from demonstration, approaches to learning multiple task variants, and active learning of control strategies based on uncertainty or confidence scores.

\subsection{Active Learning From Demonstrations}
LfD makes learning complex control applications~\cite{LFDSurvey, abbeel2010autonomous, calinon2009robot, finn2017one} easy but demonstrations are expensive. Active learning aims to ameliorate this by seeking to select the most informative learning experiences~\cite{ALSurvey}. We will show in subsection~\ref{subsection:devaition_as_proxy_for_episodic_reward} that our work leverages the predictive uncertainty as a cheap proxy as proposed in~\cite{DBLP:journals/corr/AmodeiOSCSM16} to gauge what is informative. The work in \cite{doggedLearning}, \cite{tellexAskForHelp} and \cite{chernovaCBA} quantify their uncertainty and detect unfamiliar and ambiguous states through an additional mechanism, e.g. depending on formulated pre- and postconditions~\cite{tellexAskForHelp}. Such an approach can be hard to scale up. Our proposed approach uses a single neural network that both controls the system and also detects unfamiliar  states that makes it more scalable.

\subsection{Learning Task Variants}
Subtle changes between train-time and test-time conditions might cause controllers to fail silently and unpredictably~\cite{DBLP:journals/corr/AmodeiOSCSM16} as highlighted in semi-supervised learning under domain shift~\cite{zhu2005semi}, robust generalization~\cite{zhang2016understanding}, multi-task learning~\cite{ruder2017overview} and transfer learning~\cite{pan2010survey}. The different conditions can thus be modelled as different (but closely related) \emph{tasks}. We describe such different conditions by the term \emph{contexts}.

Transfer learning addresses how policies can be transferred positively~\cite{pan2010survey, torralba2011unbiased} when there are subtle changes between train-time and test-time conditions which might cause controllers to fail silently~\cite{DBLP:journals/corr/AmodeiOSCSM16, zhu2005semi, zhang2016understanding, pan2010survey}. There have been attempts like allowing the simulator policy to initialize the real world policy~\cite{rusu2016sim}, to allow policy adjustment model that compensates for simulator-real world discrepancies~\cite{higuera2017adapting}, training a RNN across multiple MDPs, to obtain useful biases that will aid the learning of other tasks in the future~\cite{wang2016learning, RL2}. However, these approaches lack an easily scalable way to detect training-testing distribution discrepancies and quantify confidence in a transfer.

\subsection{Leveraging Uncertainty for Active Learning of Control}
PILCO~\cite{deisenroth2011pilco} uses Gaussian processes for learning a probabilistic dynamics model and long-term planning. However, Gaussian processes with generic kernels typically do not work well with high-dimensional inputs. In Deep-PILCO~\cite{gal2016improving, gamboa_nn_deep_pilco_17} a dropout training approach~\cite{MCDropout} was used to quantify the uncertainty of a learned neural network dynamics model. As an alternative, a heuristic measure has been used to quantify trust on the demonstrations in~\cite{CHAT}. However, unlike in our work, all these methods focused on a single task, and most of them focus on model-uncertainty and not policy-uncertainty.

\subsection{Learning sequential information}
There are four broad classes of methods for capturing sequential information. The easiest to use with any off-the-shelf supervised learning methods, which we also employ in our work is using a \emph{temporal windows} on sequences~\cite{frank2001time, gashler2014training}. The second method is using \emph{recurrent models} such as LSTMs which are relatively difficult to train and work with evenly sampled sequences~\cite{nerrand1994training}. The third technique which works well with unevenly sampled data is \emph{regression-based extrapolation} where training data is fitted to a curve~\cite{drucker1997support, godfrey2018neural}. Lately, \emph{Convolution based NN} techniques have been shown to perform at least as good as recurrent models for learning from sequences~\cite{bai2018empirical}, \cite{van2016wavenet}, \cite{kalchbrenner2014convolutional}, \cite{dauphin2016language}, \cite{gehring2016convolutional}.

\section{BACKGROUND}\label{section:Background}
The methods we describe in this paper require controllers $\pi$ that provide a measure of uncertainty for their outputs, which can be used by an agent to assess its confidence about applying an action at a given state. Ideally, we would like a measure of uncertainty that is data-dependent; i.e. uncertainty should be higher for states ${\bf s}$ where the agent has no data. This can be obtained by applying Bayesian inference. Given a dataset of demonstrations $\mathcal{D}=\lbrace(\vec{s}_i, \pi_d^*(\vec{s}_i))\rbrace$, we would like to estimate the posterior distribution over policies $p(\pi|\mathcal{D})$ to make predictions at possibly unseen states $\vec{s}_*$. This posterior distribution induces uncertainty on predictions as
\begin{equation}
    p(\vec{a}_*|\vec{s}_*) = \int p(\vec{a}_*|\vec{s}_*,\pi)p(\pi|\mathcal{D})d\pi.
    \label{eq:predictive_distribution}
\end{equation}
Two state-of-the-art methods that follow this approach are Gaussian Process (GP) regression and Bayesian Neural Networks (BNN). In this work, we will compare GP regression with a specific method for BNNs: Bayes-by-Backprop.

\subsection{Gaussian Process Regression}\label{Background:gaussian_process}
Gaussian Processes are a popular \emph{non-parametric} method for solving regression problems like the one in equation~\ref{eq:predictive_distribution}. GP regression proceeds by assuming that different evaluations of $\pi(\vec{s})$ for $\vec{s} \in \mathcal{D}$ are jointly Gaussian. This makes evaluating the integral in Eq.~\ref{eq:predictive_distribution} tractable. GP regression requires specifying a \emph{mean function} $m(\vec{s})$ and a \emph{kernel function} $k(\vec{s},\vec{s}')$. The mean functions acts as a prior for $\pi$, while the kernel function models the covariance of the outputs $\pi(\vec{s}), \pi(\vec{s'})$ given the inputs $\vec{s},\vec{s}'$. In this case, the parameters of the policy $\vec{\theta}$ correspond to the hyper-parameters of $m(\vec{s})$ and $k(\vec{s},\vec{s}')$. Fitting is done by optimizing $\vec{\theta}$ to maximize the marginal likelihood $p(\vec{a}|\vec{s})$ for all $\vec{s}, \vec{a} \in \mathcal{D}$.

Since the outputs are assumed to be jointly Gaussian, the resulting predictive distribution $p(\vec{a}_*|\vec{s}_*)$ is a Normal distribution $\mathcal{N}(\vec{a}_*|\vec{\mu}(\vec{s}_*), \vec{\sigma}^2(\vec{s}_*)\vec{I})$, where the parameters $\vec{\mu}(\vec{s}_*), \vec{\sigma}^2(\vec{s}_*)$ are calculated exactly by integrating~(\ref{eq:predictive_distribution}) analytically. We use $\vec{\mu}(\vec{s}_*)$ as the action to be applied by the agent and $\vec{\sigma}^2(\vec{s}_*)$ as a measure of its uncertainty. Although inference can be done exactly, this method has a higher computational cost than inference with BNNs. We refer the reader to~\cite{rasmussen2004gaussian} for details on how to compute these quantities.

\subsection{Bayes-by-Backprop}\label{Background:BBB}
Bayes-by-Backprop~\cite[BBB]{BayesByBackprop} is an approximate variational inference method for training BNNs. The posterior distribution in this case is defined over the parameters of the model, $P(\pi|\mathcal{D}) = P(\vec{w}|\mathcal{D})$, where $\vec{w}$ are the weights of the neural network model. Using the true posterior distribution for making predictions with neural networks is generally intractable. Approximate variational inference algorithms like BBB introduce an approximate posterior $q_{\vec{\theta}}(\vec{w})$ which is easy to evaluate, and predictions are done via Monte Carlo sampling:
\begin{equation}
    p(\vec{a}_*|\vec{s}_*) \approx \frac{1}{M}\sum_{i=1}^{M} p(\vec{a}_*|\vec{s}_*,\pi_{\vec{w}_i}), \quad \vec{w}_i \sim q_{\vec{\theta}}(\vec{w})
    \label{eq:bbb_predictive_distribution}
\end{equation}
Fitting the model is done by minimizing the Kullback-Leibler divergence between the true and approximate posterior, which may be expressed with an objective of the form
\begin{equation}\label{equation:tractable_cost_function}
\mathcal{L}(\mathcal{D},\theta) = -\mathbb{E}_{q_{\vec{\theta}}(\vec{w})}\left[ p(\mathcal{D}|\vec{w}) \right] + \mathrm{KL}\left(q_{\vec{\theta}}(\vec{w}) | p(\vec{w})\right)
\end{equation}
where $p(\mathcal{D}|\vec{w})$ is the likelihood of the data given the model, and $p(\vec{w})$ is a user-selected prior distribution over the weights. By making use of the reparameterization-trick~\cite{kingma2015variational,BayesByBackprop}, the objective (\ref{equation:tractable_cost_function}) can be approximated with Monte Carlo samples. This can be optimized via stochastic gradient descent with gradients evaluated by back-propagation. BBB sets $q_{\vec{\theta}}(\vec{w})$ as a Gaussian distribution with diagonal covariance $q_{\vec{\theta}}(\vec{w}) = \mathcal{N}(\vec{w}|\vec{\mu}, \vec{\sigma}) = \prod_{w_j\in \vec{w}} \mathcal{N}(w_j|\mu_j, \sigma_j)$. A suitable prior distribution in this case is a zero-mean Gaussian with diagonal unit covariance; $p(\vec{w}) = \mathcal{N}(\vec{0}, \vec{\mathrm{I}})$.

Once the parameters $\vec{\theta}^*$ that minimize (\ref{equation:tractable_cost_function}) have been obtained, predictions are done via equation~(\ref{eq:bbb_predictive_distribution}). This can be done by approximating $p(\vec{a}_*|\vec{s}_*)$ as a Normal distribution $\mathcal{N}(\vec{a}_*|\vec{\mu}(\vec{s}_*), \vec{\sigma}^2(\vec{s}_*)\vec{I})$, where the parameters $\vec{\mu}(\vec{s}_*), \vec{\sigma}(\vec{s}_*)$ are estimated empirically from multiple evaluations of the model $\pi_{\vec{w}_i}(s_*)$, with $\vec{w}_i \sim q_{\vec{\theta}^*}(\vec{w})$. Again, the predictive mean $\vec{\mu}(\vec{s}_*)$ is used as the action applied by the agent, while the predictive variance $\vec{\sigma}^2(\vec{s}_*)$ provides a data-dependent measure of uncertainty.

\section{LEARNING FROM DEMONSTRATIONS WITH BAYESIAN NEURAL NETWORKS ON MULTIPLE CONTEXTS}\label{section:the_proposed_mechanism}
In this section, we first explain how we quantify uncertainty. We then build on this to perform active learning on multiple contexts in the LfD framework. In short, we use the predictive standard deviation of BBB to quantify the confidence in the controller performing well on the current context. High uncertainty means low confidence that following the policy will be successful. 

To detect non-generalizable contexts, we use BBB to learn policies for which the inputs are temporal windows of the interaction history~(see  subsection~\ref{subsection:moving_window}). The associated uncertainty is used to  decide whether the agent is confident enough that it can perform well in the current context, or if it should request demonstrations for additional training on the current context~(subsection~\ref{subsection:low_pass_filter}).

The novelty in our mechanism lies in  three parts that interact with each other, and together allow the application of Bayesian methods to quantify uncertainty in LfD policies: temporal-windows for input representation, learning policies for generating both actions for control and uncertainty over a given context, and the detector that decides whether the current policy generalizes well to the current context. These three elements are described in detail in the sections below, and summarized in Algorithm~\ref{algorithm:proposed_mechanism}.

\subsection{Temporal Windows}\label{subsection:moving_window}
In order to implicitly identify differences in contexts and adapt to those changes, we use short histories of $k$ time-steps  or \emph{temporal windows} instead of just a single state as input to the policy. For example, if the context dynamics depends on the mass of a system, the greater inertia can only be identified when comparing two subsequent states while a known force is applied.

The dataset $\mathcal{D}$ for training is extracted from the demonstrations with such temporal windows, with a stride of one time-step. The overlapping windows $\vec{x}_t$ consist of the last $k$ states $\vec{s}_{t-k:t}$ paired with their associated actions $\vec{a}_{t-k:t-1}$ and rewards $r_{t-k:t-1}$.

\subsection{Learning BNN Policies}\label{subsection:learning_policies}
The dataset is created as described in subsection~\ref{subsection:moving_window} and the BBB is trained by minimizing Eq.~\ref{equation:tractable_cost_function}. Hence, the parameters $\vec{\theta}^*$ of the learned controller $\hat{\pi}$ are defined as
\begin{equation*}\label{equation:training_policy}
    \begin{split}
        &\vec{\theta}^* = \argmin_{\vec{\theta}}\mathcal{L}(\mathcal{D},\vec{\theta}).
    \end{split}
\end{equation*}
After optimization, the controller computes its prediction of the expert action and its uncertainty about the current situation with $\vec{x}_t$ as input,  as:
\begin{equation*}
        \vec{a}_t, \vec{\sigma}_t = \Hat{\pi}_{\vec{\theta}^*}(\vec{x}_t),
        \textnormal{ with } 
        \vec{a}_t = \mathbb{E}\left[\hat{\vec{y}}|\vec{x}_t\right], \,
        \vec{\sigma}_t = \sqrt{\operatorname{Var}\left[\Hat{\vec{y}}|\vec{x}_t\right]}. \\
\end{equation*}
which are estimated using Monte-Carlo samples. The vector $\vec{\sigma}_t$ is then converted to a scalar $\sigma_t$ by taking the average over the standard deviations obtained for each action-dimension.

\subsection{Detector}\label{subsection:low_pass_filter}
The robot will encounter situations with contexts across the whole spectrum between being similar enough for good generalization to being quite different. Based on the controller's confidence, the agent decides whether to continue, or to stop and ask for demonstrations of the current context.

Particularly, we first smooth $\sigma_t$ with a moving average over $m$ time steps. The controller deems itself to be not confident enough to proceed if the average uncertainty exceeds a scaled adaptive threshold $c\Omega$ where $c$ is a scaling factor and $\Omega$ is the adaptive threshold. The value of adaptive threshold~$\Omega$ is automatically set to the average uncertainty obtained over all the contexts the controller has trained itself on so far. Both the constants $c$ and $m$  can be  tuned to yield more or less sensitive learners, which we demonstrate in subsection~\ref{subsection:tuning_conservativeness}.

\subsection{Connecting the dots} \label{subsection:joining_all_pieces}

\begin{algorithm}
\caption{Select action and train if necessary}
\label{algorithm:proposed_mechanism}
\begin{algorithmic}[1]
        \REQUIRE $\vec{x}_t, \mathcal{D}$,$\vec{\theta},d,c$
        \STATE $\vec{a}_t, \sigma_t \leftarrow \Hat{\pi}_{\vec{\theta}}(\vec{x}_t)$
        \IF{smoothed($\sigma_t$) $\!>\!c\Omega \,\textbf{and}\, t \!>\! t_{start}$}
            \STATE $\mathcal{D} \leftarrow \mathcal{D} \cup  \text{GetDemonstrations($d$)}$
            \STATE $\vec{\theta}^* \leftarrow \argmin_{\vec{\theta}}\mathcal{L}(\mathcal{D},\vec{\theta})$ 
            \hfill\#relearn
            \STATE $t\leftarrow 0$
            \COMMENT{re-start episode in current context}
            \STATE $\Omega = \text{AverageUncertainty()}$ \COMMENT {adapt query threshold}
            \STATE $\vec{a}_t, \sigma_t = \Hat{\pi}_{\vec{\theta}^*}(\vec{x}_t)$
            \COMMENT{re-select action}
        \ENDIF
        \RETURN $\vec{a}_t,\mathcal{D},\vec{\theta}^*$
\end{algorithmic}
\end{algorithm}

\begin{figure}
    \centering
    \includegraphics[width=\columnwidth]{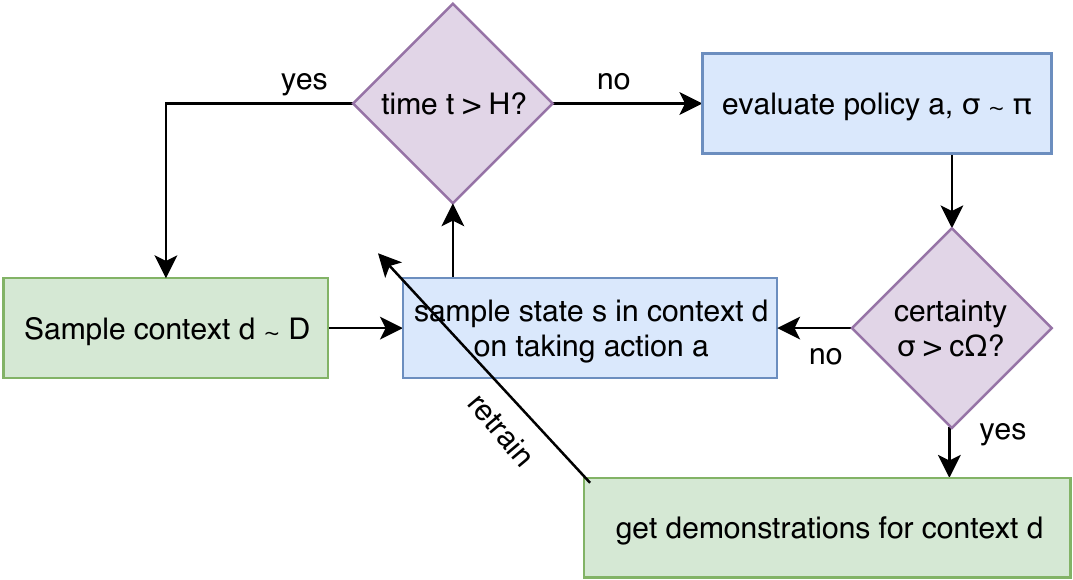}
    \caption{The controller has to perform well on all tasks it faces sequentially with limited requests for task-specific demonstrations. A certainty threshold on the predictive variance is used to decide when to ask for demonstrations.}
    \label{fig:problem_set_up}
\end{figure}

Algorithm~\ref{algorithm:proposed_mechanism} describes the steps of our approach, with Figure~\ref{fig:problem_set_up} showing how our method is used in an LfD setting. Using a  BNN instead of a deterministic neural network allows us to  obtain the predictive standard deviation along with the action for each temporal window. The temporal windows contain the robot's state information over  consecutive time-steps, which allows the BNN to implicitly infer the underlying dynamics. The uncertainty  is passed to a detector that smooths this signal and compares it to a scaled threshold. When the uncertainty is higher than the scaled threshold, continuing will likely lead to  poor performance. Hence, execution is paused and more training is done using new demonstrations solicited on the context at hand. The value of adaptive threshold~$\Omega$ is initialized with $0$ so that it always trains on requested demonstrations during the first context it faces. Following each training episode, the threshold $\Omega$ is adapted to the average uncertainty obtained on all the contexts that the controller has previously trained itself on. 

\section{EXPERIMENTS}\label{section:Experiments}
Our experiments\footnote{The code we used for these experiments is available at \url{https://git.io/fA1E4}} consist of training robust controllers for different task \emph{contexts}. These closely related contexts are obtained by varying variables of the environment, for example the masses and lengths of parts of the system to be controlled. These variables are not directly observable by the learning agent, but do have an effect on the outcome of applying an action. Since the learner is not told the context explicitly, it has to infer it from the past transitions within the temporal window. For all experiments, we require \emph{expert} demonstrations. While our system can in principle operate with a human demonstrator, for purposes of reproducible science, we base our experiments on previously optimized agents. We use the definitions of episodic reward~($r_d$) and standard deviation~($\sigma_d$) on context~$d$ as $r_d = \sum_{t=0}^{H} r_t$ and $\sigma_d = \sum_{t=0}^{H} \sigma_t$ in our experiments.

\subsection{Simulated environments}
We tested our method on two modified OpenAI Gym environments~\cite{brockman2016gym}: HalfCheetah and Swimmer. In both environments, the goal is to find locomotion controllers to move forward as fast as possible. To simulate different contexts, we changed either or both the lengths and masses of various body parts like torso, middle, and back. For these experiments, we used proximal policy optimization~(PPO)~\cite{PPO} to obtain expert policies for each context.

We also compare the relative learning performance of BBB to that of the GP on a simpler, low-dimensional task: a double integrator domain of a point mass sliding on a friction-less surface, where the goal is to apply forces  so as to come to an halt exactly at coordinate $x=0$. The demonstrator here is an LQR based controller obtained by solving the Riccati equation.

\subsection{Physical robot experiment}
We tested parts of our method on a physical Furuta pendulum \cite{furuta}. The goal in this environment is to swing-up and balance the pendulum upright, by applying a torque to the base joint. This is analogous to the standard cart-pole swing-up task, with $4$ continuous state dimensions and $1$ continuous action dimension. We varied the pole mass to create different contexts. This was done by attaching weights of varying sizes to the pole. We used the model-based method described in~\cite{gamboa_nn_deep_pilco_17} to obtain expert policies for each context.

\begin{figure}
    \centering
    \includegraphics[width=\linewidth]{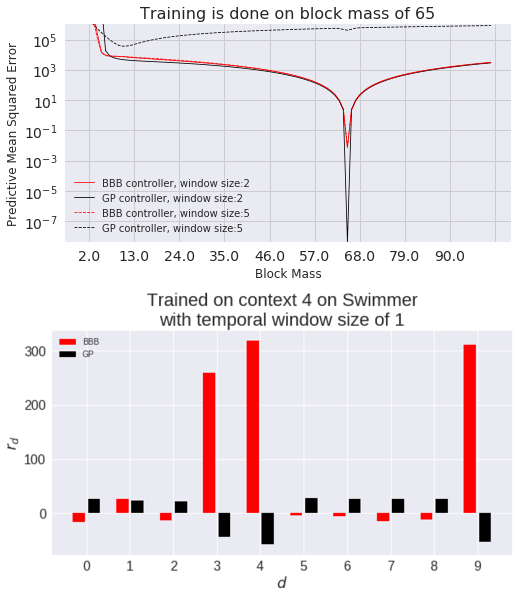}
    \caption{GPs do not do as well as BBB when the number of input dimensions increases. Top figure shows how the performance of GP deteriorates with the increase in dimensions where the number of input dimensions  corresponding to window sizes of $2$ and $5$ is $6$ and $18$ respectively. As a result, they never work with tasks that are already high dimensional (bottom figure) even with a temporal window size of $1$, let alone use of deeper histories.}
    \label{fig:BBBvsGP}
\end{figure}

\section{RESULTS}\label{section:Results}
In this section, we will first show that our approach is more scalable and hence can do well even in high-dimensional tasks. We will then evaluate whether the confidence of our policy is a good proxy for success. Then, we will show how we can leverage the confidence to reduce the number of conditions in which demonstrations need to be given, and investigate the effect of the main tunable hyperparameters. 

\subsection{BNNs outperform GPs in high-dimensional tasks}\label{results:BBB_vs_GP}
We compare our BNN based proposed mechanism with GP in figure~\ref{fig:BBBvsGP}. Although GPs can do as well as BNNs in lower-dimensional domains, they fare significantly worse in higher-dimensions. In contrast, the number of input dimensions has an insignificant effect on the BNN's ability to learn. This is critical in most of the interesting real-world problems as they are high-dimensional. 
\begin{figure}
    \centering
    \subfloat[]{\label{figure:our_real_pendulum}\includegraphics[clip,width=0.55\columnwidth,height=100px]{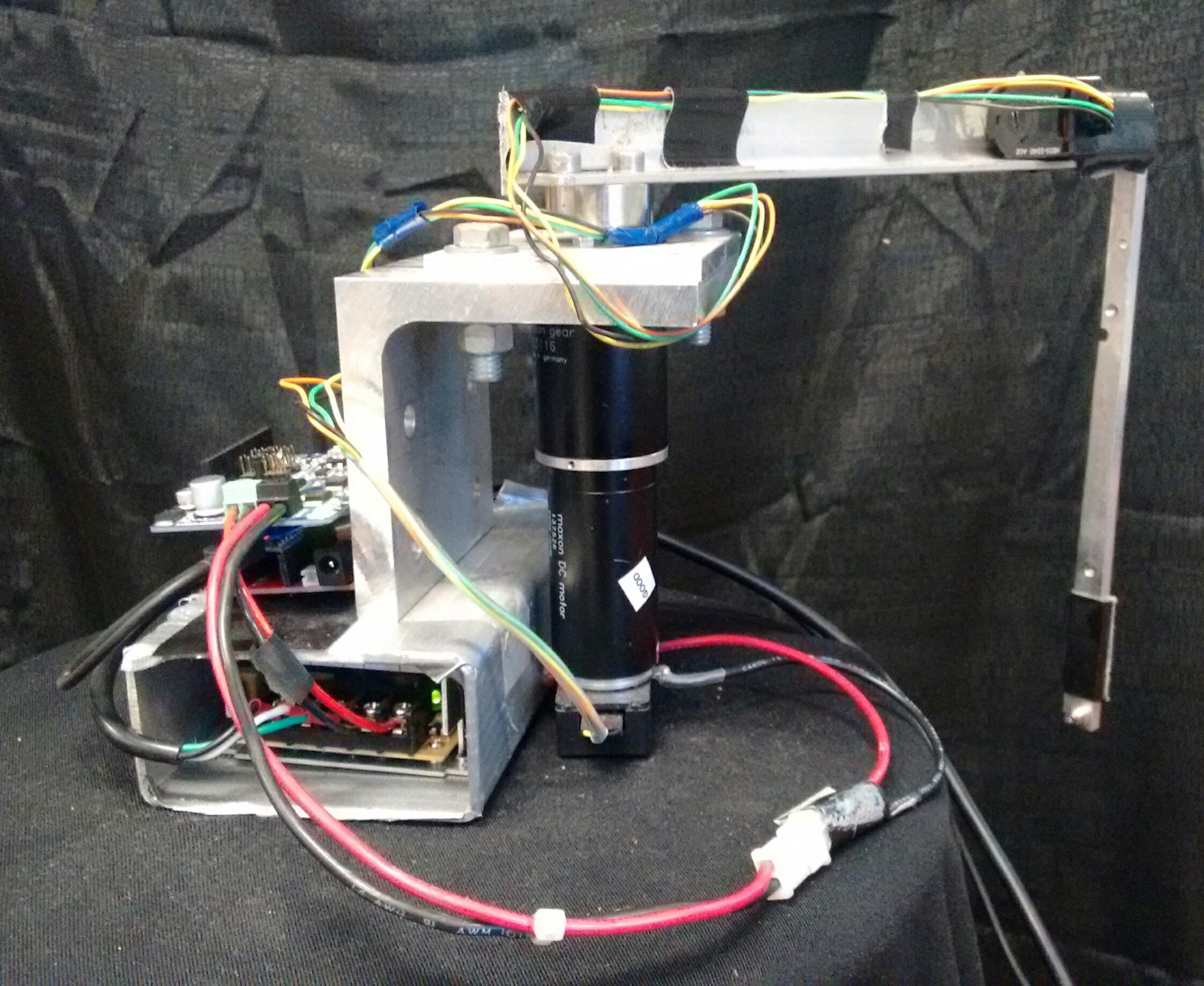}} \\
    \subfloat[]{\label{figure:reward_and_deviation}\includegraphics[clip,width=\columnwidth]{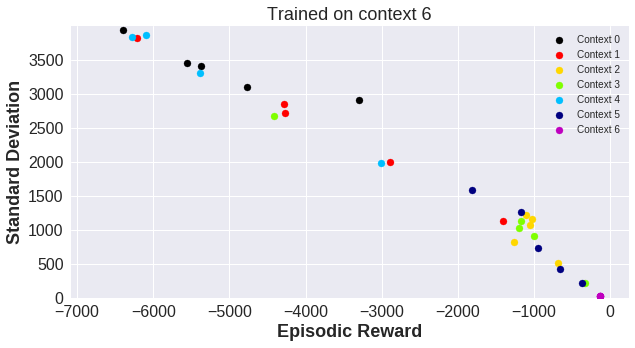}}
    \caption{(a) Illustration of the experimental set-up for the robotic experimental platform. (b) For each  swing-up task context,  $\sigma_d$ and  $r_d$ obtained during five independent runs are plotted against each other.  Note the strong relationship between these quantities: higher $\sigma_d$ is correlated with lower $r_d$. Thus, $\sigma_d$ can be used as a predictor for task success. Training was performed on task context $6$ in this case.}
    \label{fig:real_pendulum}
\end{figure}

\subsection{Uncertainty as a predictor of success}\label{subsection:devaition_as_proxy_for_episodic_reward}
In our work, the predictive uncertainty $\sigma_d$ is used to detect and possibly preempt the execution on contexts the policy is unlikely to do well on.
 To test whether $\sigma_d$ is indeed a good predictor for success, we compared the obtained $\sigma_d$ and its corresponding empirical episodic reward $r_d$. We performed this experiment on the physical Furuta pendulum. Although the imitation learner does not attempt to optimize the reward, the demonstration policy does. Thus, high rewards are indicative of the learner closely following the demonstrator. 
 
 Figure~\ref{figure:reward_and_deviation} shows that $\sigma_d$ is indeed related directly to the performance measurement $r_d$, so it can be used to predict whether the learner will do well. In particular, if the learner has been trained on contexts that are similar to the new context, it will tend to have a low predictive uncertainty and generalize well. If the new context is quite different, the policy will tend to have a high predictive uncertainty, and instead of attempting to perform the new context, the learner will first solicit new demonstrations for this context and then re-trains its policy on the full data-set.

\subsection{Data efficiency by deciding when to query}
One way of validating if our proposed active learner invokes the fallback strategy effectively is by comparing the number of requests it makes for context-specific demonstrations with a \emph{naive baseline} that solicits such context-specific demonstrations on every context it faces and a \emph{random baseline} that never seeks any demonstration. We do two kinds of analysis for evaluating our active learner:
\begin{enumerate}
    \item \textbf{Active LfD using predictive uncertainty sanity check: }Can the learner effectively ground its adaptive threshold in tandem with generating predictive uncertainty in any situation in order to correctly predict whether the controller can generalize to that context?
    \item \textbf{Data-efficiency and tuning learner conservativeness:} How judiciously the learner seeks demonstration and how does it affect the overall performance with respect to the \emph{naive} and \emph{random} baselines?
\end{enumerate}

Answering these questions entails gathering data and evaluating policies over multiple  task contexts. To keep experiments doable, we will do so in simulation rather than on the real robot system. 

\subsubsection{Active LfD using predictive uncertainty sanity check}
\begin{figure}
\centering
\includegraphics[width=0.9\linewidth]{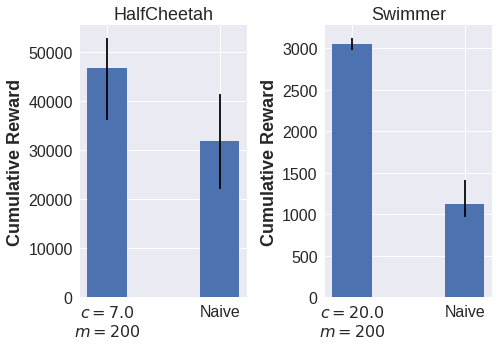}
  \caption{Cumulative reward over all contexts in $D$ after making $2$ requests for demonstrations. Our active learner outperforms the naive learner that requests demonstrations in every context. Bars show average rewards over $10$ episodes with error bars showing minimum and maximum values.}
  \label{figure:main_result}
\end{figure}

We first do a sanity check to see if our active learner has the ability to adapt its threshold to actively request context-specific demonstrations when it is most necessary. For this we purposefully arrange the sequence of contexts in a way that the first two contexts are similar and different from the third. While the first two demonstration request for \emph{naive} controller would be on the first and second context, our active learner's first two requests should correspond to the first and the third context. Figure~\ref{figure:main_result} shows that our proposed active learner achieves higher cumulative rewards over all the contexts in $D$ than the \emph{naive} controller after $2$ requests for demonstrations which is a reflection of the fact that our active learner learns to perform well across a larger number of contexts than a naive controller with the same number of requests for demonstrations. The exact details of the ordering of the task can be found in the provided code.

\subsubsection{Data-efficiency and tuning learner conservativeness}\label{subsection:tuning_conservativeness}
\begin{figure}
    \centering
    \includegraphics[width=1.0\linewidth =82px]{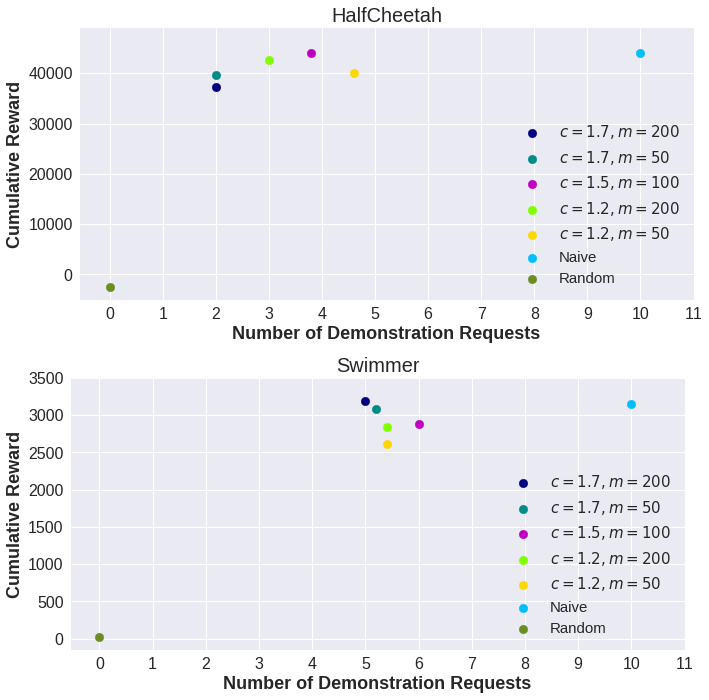}
    \caption{The plots above show that lower $c$ and $m$ leads to seeking more context-specific demonstrations or more conservative behavior without much gain in reward. The plots also show that performance close to a \emph{naive} learner that seeks demonstrations on every context, can be obtained by lesser number of such requests. The results were obtained by averaging over $5$ simulation runs with different randomized orderings of contexts in $D$.}
    \label{figure:tuning_conservativeness}
\end{figure}

After validating the potential of our active learner to make informed decisions about where to request for demonstrations, we now randomize the ordering of contexts in $D$ to analyze data requirements for successfully solving all contexts in $D$. Figure~\ref{figure:tuning_conservativeness} shows various configurations of our active learner using combinations of $c$ and $m$ to solve all $D$ contexts. With a moderate number of requests for demonstrations, our proposed agents learn to solve the task adequately - appreciably better than \emph{random} and close to the \emph{naive} learner. hence attaining data-efficiency (minimizing the number of, possibly expensive, demonstrations). Note that the \emph{naive} learner here solicits demonstrations on all the contexts it faces.

We would also like to note the effect of hyperparameters $c$ (threshold scale) and $m$ (amount of smoothing) on the conservativeness (asking for more demonstrations in case of doubt). Note that lower $c$  and lower $m$ leads to the agent asking for more demonstrations. These additional demonstrations offer greatly diminished returns, as the average reward attained did not go up appreciably as a result.

\section{DISCUSSION AND FUTURE WORK}\label{section:Future_Work}
In this work, we proposed a mechanism based on Bayesian Neural Networks that can detect when task dynamics have changed significantly from the training dynamics. In a robotic pendulum swing-up task, we showed that the confidence of the learned controller obtained by our method is indeed a good predictor of success. Furthermore, as a proof of concept of effectively plugging in a fallback strategy, we showed how this mechanism can be used to judiciously request for demonstrations in novel conditions. The proposed Bayesian Neural Network scales better to a higher-dimensional task than a Gaussian process baseline.

We expect that our mechanism for detecting lack of confidence can be used to increase safety in robotics tasks. Depending on the task, our BNN controller could be combined with other fallback strategies such as a back-up controller or  gracefully coming to a stop.

An avenue of future research is to combine our approach with dataset aggregation~\cite{Dagger} to deal with the inherent LfD problem of covariate shift between the demonstrator's and the learner's state distribution. Furthermore, it is not always clear what aspects of a demonstration are critical for the desired behavior. We will investigate how to integrate a mechanism such as the one described by~\cite{dorsa2017active} to  actively query a human supervisor and address this issue.

\section*{ACKNOWLEDGMENTS}

We gratefully acknowledge the Natural Sciences and Engineering Research Council of Canada~(NSERC) for funding and NVIDIA Corporation for the GPU donation.

\clearpage


\end{document}